\documentclass[11pt]{article}

\usepackage{microtype}
\usepackage{graphicx}
\usepackage{booktabs} 
\usepackage{caption}
\usepackage{subcaption}
\usepackage[numbers,sort&compress]{natbib}
\usepackage{lineno}
\usepackage{color}

\usepackage{hyperref}
\usepackage{chemformula}

\usepackage[margin=2.5cm]{geometry}

\begin{document}

\title{Online learning to accelerate nonlinear PDE solvers: applied to multiphase porous media flow}

\author{Vinicius L S Silva$^{*,a,b,c}$ \and Pablo Salinas$^{e}$ \and Claire E Heaney$^{a,d}$  \and Matthew D Jackson$^{b}$ \and Christopher C Pain$^{a,d}$}

\maketitle

\begin{center}
\small \noindent $^a$Applied Modelling \& Computation Group, Imperial College London, UK\\
\small \noindent $^b$Novel Reservoir Modelling and Simulation Group, Imperial College London, UK\\
\small \noindent $^c$Petroleo Brasileiro S.A. (Petrobras), Rio de Janeiro, Brazil \\
\small \noindent $^d$Centre for AI Physics Modelling, Imperial-X, Imperial College London, UK\\
\small \noindent $^e$OpenGoSim, Leicester, UK \\
\small \noindent $^*$ Corresponding author email: v.santos-silva19@imperial.ac.uk
\end{center}

\begin{abstract}

We propose a novel type of nonlinear solver acceleration for systems of nonlinear partial differential equations (PDEs) that is based on online/adaptive learning. It is applied in the context of multiphase flow in porous media. The proposed method rely on four pillars: (i) dimensionless numbers as input parameters for the machine learning model, (ii) simplified numerical model (two-dimensional) for the offline training, (iii) dynamic control of a nonlinear solver tuning parameter (numerical relaxation), (iv) and online learning for real-time improvement of the machine learning model. This strategy decreases the number of nonlinear iterations by dynamically modifying a single global parameter, the relaxation factor, and by adaptively learning the attributes of each numerical model on-the-run. 
Furthermore, this work performs a sensitivity study in the dimensionless parameters (machine learning features), assess the efficacy of various machine learning models, demonstrate a decrease in nonlinear iterations using our method in more intricate, realistic three-dimensional models, and fully couple a machine learning model into an open-source multiphase flow simulator achieving up to 85\% reduction in computational time. 
\\
\\
\textbf{Keywords:} 	nonlinear PDE solver, machine learning, online learning, numerical relaxation,
multiphase flows, porous media

\end{abstract}

\textbf{Article Highlights}
\begin{itemize}
    \item An online/adaptive learning acceleration for nonlinear PDE solvers applied to multiphase porous media flow is proposed.
    
    \item Various machine learning strategies and set of features are evaluated to determine their suitability to accelerate the solver.
    
    \item We fully integrate a machine learning model into the nonlinear solver achieving up to 85\% reduction in computational time. 
\end{itemize}

\section{Introduction}\label{sec:intro}

The numerical solution of partial differential equations (PDEs) is an ubiquitous tool for modelling physical phenomena. Among PDE problems, multiphase flow in porous media is crucial for understanding, developing and managing subsurface reservoirs, with applications in areas such as carbon sequestration, hydrocarbon recovery, geothermal energy extraction, and groundwater resources. However, the intricate coupling between the governing equations and their inherent nonlinearity present substantial challenges to the numerical solution \citep{aziz:79,jackson:15,li:15,pour2023nonlinear}. Traditionally, Newton methods in monolithic fully-implicit formulations have been employed to solve these discretized nonlinear equations; however, sequential methods are emerging as a competitive alternative due to their flexibility and extensibility. Sequential methods allow the use of specialized solvers for the solution of each equation in multiphysics problems \citep{salinas:17, jiang:19, freitas:20}. The Picard iterative solver can function as a sequential method, which can be conceptualized as a nonlinear block Gauss-Seidel problem or a sequential fixed-point iteration \citep{silva:21machine}. Compared to Newton-like methods, Picard iterative solvers have less stringent convergence criteria \citep{elman:05, lott:12}. Newton methods can also be used in sequential approaches within both inner and outer loops. Their quadratic convergence makes them generally faster than Picard iterations \citep{Ortega:70, jenny:06, wong:19}.

Significant research has been performed to accelerate sequential methods. To this end, the use of numerical relaxation \citep{press:07numerical} has shown promising results. \citet{salinas:17} introduced an acceleration technique employing numerical relaxation, targeting the saturation field in multiphase porous media problems. \citet{jiang:19} implemented three distinct nonlinear acceleration techniques in the sequential-implicit fixed point method, with numerical relaxation outperforming both quasi-Newton and Anderson acceleration.

The widespread success of machine learning across various domains has recently motivated its application in accelerating the convergence of numerical PDE simulations.
\citet{greenfeld:19} proposed a new framework for multigrid PDE solvers, where a single mapping from discretization matrices to prolongation operators is learned using a neural network in an unsupervised learning procedure. \citet{hsieh:19} proposed an approach to learn a fast iterative PDE solver tailored to a specific domain. It achieved significant speedups compared to standard formulations; however, it only works for linear solvers. 
\citet{oladokun:20} implemented a random forest regression to generate the linear convergence tolerance in a nonlinear solver, resulting in a reduction of linear iterations.
\citet{kadeethum2022enhancing} used the prediction from a machine learning-based reduced-order model as a initial guess for the nonlinear solver iterations. Their approach maintains the full order model accuracy while accelerating the nonlinear solver convergence. 
In a previous work \citep{silva:21machine}, we proposed a machine learning acceleration to determine the value of the numerical relaxation. We used a random forest model (trained offline) and a selection of dimensionless parameters for the training and controlling the machine learning algorithm. The machine learning acceleration was able to generalise, and improve the stability and efficiency of the nonlinear solver.

A promising approach in the field of machine learning comes in the form of learning from online/evolving data streams \citep{read2012batch,kirkpatrick2017overcoming,gomes2017adaptive,chen2018lifelong,montiel2020adaptive,hoi2021online}. It can provide an attractive alternative to accelerate nonlinear PDE solvers, since in each numerical simulation several nonlinear iterations are performed, and they can be used to improve the convergence of the nonlinear solver. 
We can highlight two main branches of algorithms to train the machine learning model, instance-incremental methods \citep{cauwenberghs2000incremental,oza2001online,losing2016knn,hoi2021online} and batch-incremental methods \citep{breiman1999pasting,polikar2001learn++,wang2003mining,montiel2020adaptive}. Instance-incremental learning uses only one sample to update the machine learning model at each time, where batch-incremental learning uses a batch of multiple samples to update the model each time. \citet{read2012batch} compared these two approaches for the task of classification of evolving data streams. They showed that both methods perform similarly given a limited resource, and that the optimal batch size depends on the problem in consideration.

In this work, as a starting point, we perform a sensitivity in the machine learning models and features of the machine learning acceleration proposed in our previews work \citep{silva:21machine}. Then we use an online learning approach to improve/update the machine learning model in run time, with the aim of continuously adapting to changes in the numerical PDE simulation. To the best of the authors' knowledge, this is the first work to use online learning to accelerate a numerical PDE solver.  

We list the contributions of our work as follows:
\vspace{-1ex}
\begin{itemize}
\itemsep0em 
    \item We propose an online/adaptive learning acceleration for nonlinear PDE solvers, and apply it to a broad class of multiphase porous media problems. It is able to control the convergence of the nonlinear solver by learning from previous nonlinear iterations of the running numerical simulation.
    \item We investigate the best set of features (dimensionless numbers) used as input parameters for the proposed acceleration and compare different machine learning models for controlling the convergence of the solver.
    \item We fully integrate a machine learning model into the nonlinear solver of an open-source multiphase flow simulator, achieving up to 85\% reduction in computational time (only using the offline training, the online update would further improve the results).
    \item The proposed methodology can be applied to other types of nonlinear solvers provided they use a numerical relaxation or another tuning parameters that influences the solver convergence.
\end{itemize}
\vspace{-1ex}

The remaining of this paper is structured as follows. We present the governing equations of the multiphase porous media flow and the Picard iterative solver in Section \ref{sec:back}. Section \ref{sec:method} describes the proposed adaptive learning acceleration. Following that, the test cases are detailed in Section \ref{sec:apptc} and the results are presented in Section \ref{sec:exp}. Finally, a discussion and concluding remarks are provided in Section~\ref{sec:disc} and \ref{sec:conc}, respectively. 

\section{Background and problem setting}\label{sec:back}

In this section, we show the governing equations of the multiphase porous media flow problem. Then we describe the Picard iterative solver and the numerical relaxation used here. 

\subsection{Governing equations and discretization}

We report the governing equations for incompressible porous-media flow considering capillary pressure and gravity. The mass-balance equation is 
\begin{equation}\label{eq:massbal}
\phi \frac{\partial S_\alpha}{\partial t} + \nabla\cdot \mathbf{u_\alpha} = Q_\alpha,
\end{equation}  
where $\alpha$ represents an immiscible-fluid phase and  $t$ is time. $\phi$ is the rock porosity, $\mathbf{u}$ is the Darcy velocity, $S$ is the saturation, and $Q_\alpha$ is a source term.   

The relation between pressure and the Darcy velocity is given by the  multiphase Darcy's law for phase $\alpha$ as
\begin{equation}\label{eq:darcy}
\mathbf{u_\alpha} = \frac{k_{r\alpha} \mathbf{K}}{\mu_\alpha} (-\nabla p_\alpha + \rho_\alpha g \nabla z ),  
\end{equation}
where $\mu_\alpha$, $k_{r\alpha}$, and $\mathbf{K}$ are the viscosity, relative permeability, and  permeability tensor, respectively. $\rho$ is the density, $p$ is the pressure, $g$ is the gravitational acceleration, and $\nabla z$ is the gravity direction. 


Considering a wetting ($w$) and non-wetting ($nw$) phase and including capillary pressure, $p_c$, the system of equations is completed by the constraints
\begin{equation}\label{eq:pc}
p_c = p_{nw} - p_w, 
\end{equation}
\begin{equation}\label{eq:sat}
S_w + S_{nw} = 1.
\end{equation}

Our approach adopts a spatial discretization strategy using the double control volume finite element method, as elaborated in \citet{salinas:17b}. This method offers an enhancement over the traditional control volume finite element method, particularly for highly distorted meshes, as discussed in \citep{fung:92, durlofsky:93, jackson:15, gomes:17}. In this framework, velocity is discretized via finite elements, while pressure and saturation are handled through control volumes. For time discretization, we employ a $\theta$-method. Here, $\theta$ is adaptively adjusted between 0.5 (Crank-Nicolson scheme) and 1 (equivalent to implicit Euler), guided by the total variation diminishing criterion \citep{pavlidis:14}.
The solution of the equations reported here are implemented in the open-source code IC-FERST (Imperial College Finite Element Reservoir SimulaTor) \citep{gomes:17,salinas:17, salinas:17b,Obeysekara2021}. Further details about IC-FERST can be found in \url{https://multifluids.github.io/}. 

\subsection{Nonlinear solver}

\begin{figure}[!tb]
	\centering
	\includegraphics[width=0.7\textwidth]{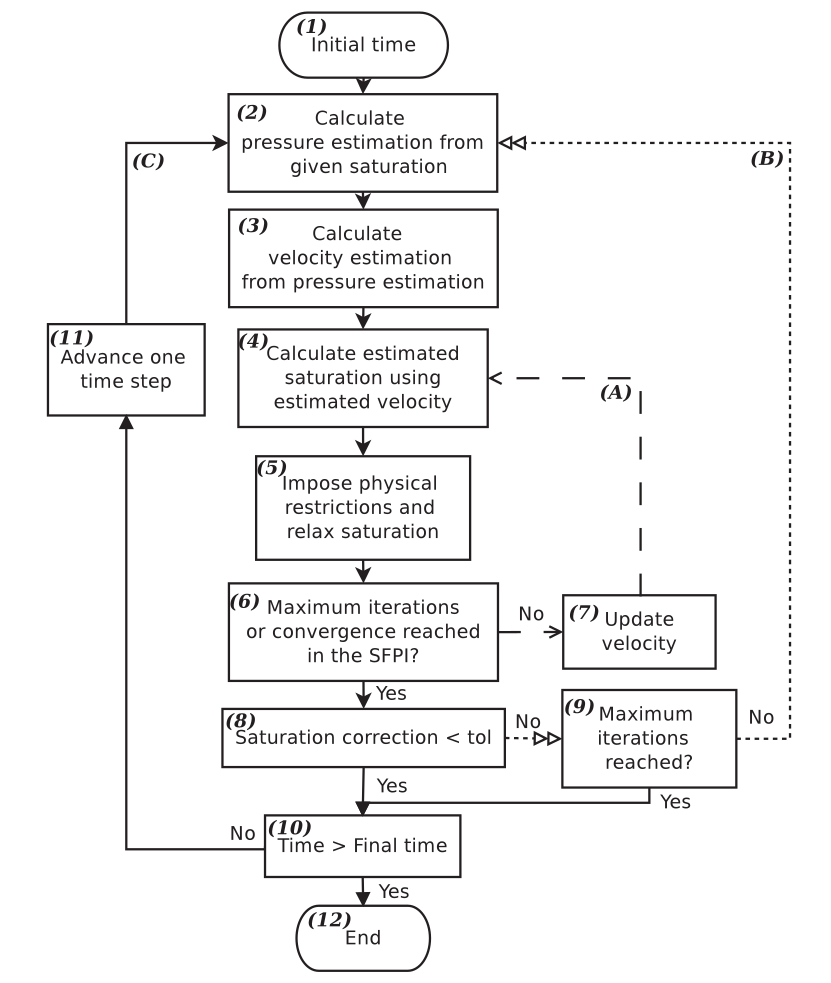}
	\caption{Sequential nonlinear solver loops. Reprinted from \citep{salinas:17}.}
	\label{fig:solver}
\end{figure}

We use a sequential method (Picard iterative method) to solve the discretized nonlinear system of equations, constituted by Eqs. (\ref{eq:massbal}), (\ref{eq:darcy}), (\ref{eq:pc}), and (\ref{eq:sat}). The method, outlined in \citet{salinas:17}, is depicted in Figure \ref{fig:solver} and involves three iterative loops. The time loop is indicated by a solid line in Figure \ref{fig:solver}, the coupling between saturation and pressure (nonlinear outer loop) by a dotted line, and the iteration over saturation and velocity (nonlinear inner loop) by a dashed line. Within the inner loop, saturation is calculated first, followed by an update in velocity based on the newly computed saturation. This cycle repeats until either convergence is achieved or the maximum number of iterations is reached. Subsequently, in the outer loop, pressure is recalculated based on the updated saturation, which in turn leads to a new velocity calculation derived from the pressure estimate. This initiates a new iteration in the inner loop. The entire process continues until either convergence is achieved or the maximum iteration limit is met.

To enhance the convergence rate of the nonlinear solver, a relaxation parameter is implemented \citep{salinas:17,jiang:19,silva:21machine}. This technique updates the saturation by integrating the newly computed field with saturations from prior inner loop iterations. The calculation for the new saturation is as follows
\begin{equation}\label{eq:relaxation}
S^k = \omega^k \tilde{S}^k + (1-\omega^k)S^{k-1} + (1-\omega^k)^{\beta+1}\omega^k(S^{k-2}-S^{k-1}), 
\end{equation}   
where $S^k$ is the updated saturation, which is calculated using the saturation $\tilde{S}^k$ from step $(4)$ in Figure~\ref{fig:solver}. $k$ denotes the current iteration of the inner loop. The relaxation parameter is represented by $\omega$, and $\beta$ is the exponent that determines the relative significance o $S^{k-1}$ and $S^{k-2}$. Here, we adopt a constant value of 0.4 for $\beta$, following the methodology outlined in \citet{salinas:17}.

In each outer nonlinear iteration, the initial relaxation parameter $w^0$ sets the stage for calculating subsequent relaxations $w^k$. The latter are computed to optimize the convergence rate by minimizing the residual saturation \citep{salinas:17}. A critical consideration is the selection of $w^0$ for each outer nonlinear iteration. While a simple approach might use a constant value across all iterations, the relaxation parameters must be carefully balanced: small enough to prevent divergence, yet large enough to expedite convergence. The choice of $w^0$ significantly influences the convergence of the nonlinear loops, but its optimal value is not predetermined and varies with each specific problem. Furthermore, even the most effective static value may lead to a higher number of iterations compared to a well-adjusted dynamic relaxation factor \citep{kuttler:08,jiang:19,silva:21machine}.

\section{Online learning acceleration}\label{sec:method}

\begin{figure}[!htb]
	\centering
	\includegraphics[width=1.0\columnwidth]{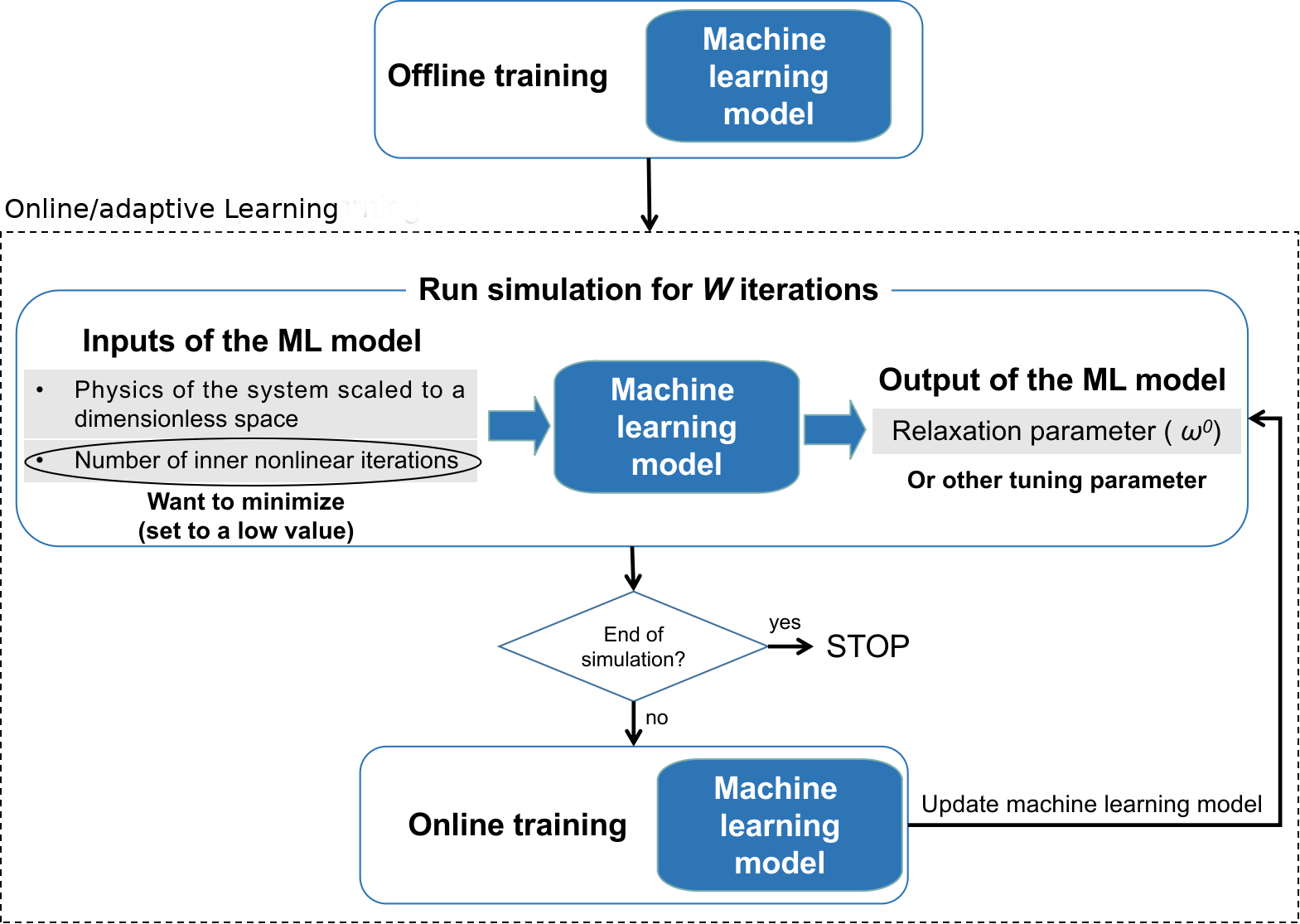}
	\caption{Adaptive learning acceleration for nonlinear PDE solvers.}
	\label{fig:mlacc1}
\end{figure}

The relaxation parameter is of paramount importance to accelerate the nonlinear solver convergence. Therefore, we propose a novel approach to adjust the relaxation factor in each outer nonlinear iteration, where we adapt the machine learning model to the changes in the numerical simulation during run time, as shown in Figure \ref{fig:mlacc1}. We start with an offline part, where the machine model is trained in a dataset generated by a simple two-dimensional reservoir model. The machine learning model is designed to take simulation properties, dimensionless numbers, and the number of inner nonlinear iterations at each outer nonlinear iteration as inputs. Its output is the calculated value of the relaxation parameter. Further details about the offline training can be found in \citet{silva:21machine}. Here, we test different sets of dimensionless numbers as input parameters, and we analyse several machine learning models in terms of accuracy, prediction time and simulation results.

After the offline stage, we use the machine learning model to determine the value of the numerical relaxation for a number of outer nonlinear iteration that we call $W$ (as in Figure \ref{fig:mlacc1}). Following that, we can update the machine learning model using a batch-incremental method  ($W>1$),  or an instance-incremental method ($W=1$). After updating, we run more $W$ outer nonlinear iterations and update the machine learning model again. The process continues until the end of the numerical simulation. Given that batch-incremental and instance-incremental methods can perform similarly \citep{read2012batch}, and to avoid the cost added to the numerical simulation each time we update the machine learning model, we choose to use batch-incremental methods here (or $W>1$). We also test different batch sizes in order to determine the best one for the solver and multiphase porous media problems used here. It is worth mentioning that since we have a fixed simulation time the number of updates of the machine learning model is finite, different from classical online learning problems. 

The main goal of the adaptive learning acceleration is to train the machine learning model offline using a simple reservoir model (that is fast to run). Then apply the proposed acceleration to more complex and challenge reservoirs, while still learning from them in run time. During the numerical simulation, we learn from previous nonlinear iterations of the running reservoir model in order to better calculate the relaxation factor for the coming iterations.


\subsection{Input parameters and dimensionless numbers}

The spatial distribution of fluids in multiphase flow through porous media are governed by the interplay of capillary, viscous, and gravitational forces \citep{hoteit:08,li:15, debbabi:17, debbabi:17b, debbabi:18b}. The relative significance of these mechanisms is influenced by a combination of system length-scales, fluid-rock properties, and flow rates \citep{debbabi:18b}. Consequently, the inputs for the machine learning model are selected to be the dimensionless numbers as outlined in \citet{debbabi:17, debbabi:17b, debbabi:18}. Employing dimensionless parameters to calculate the relaxation parameter enables the use of a simple two-dimensional layered reservoir for training, while effectively covering a broad spectrum of the physical space. All the input parameters are summarized in Table \ref{tab:inputs}. A detailed description of the feature selection is given in Section \ref{sec:feature}.    

\begin{table*}[t]
\caption{Input parameters for the machine learning model. CFL stands for Courant–Friedrichs–Lewy (CFL). The features where MIN, MAX and AVERAGE applies are defined control-volume wise.}
\label{tab:inputs}
\vskip 0.15in
\begin{center}
\begin{small}
\begin{sc}
\begin{tabular}{lcccr}
\toprule
Features & \citet{silva:21machine} & Set 1 & Set 2  \\
\midrule
Effective aspect ratio     & one value & one value & one value \\
Darcy velocity             & average, max and min & average & average \\
Total mobility             & average, max and min & average & average \\
CFL number                 & max value & max value & max value \\
Shock-front CFL number     & max value & max value & max value \\
Shock-front number ratio   & one value & one value & one value \\
Shock-front mobility ratio & average, max and min & average & average \\
Longitudinal capillary     & average, max and min & average & average \\
Transverse capillary       & average, max and min & average & $\times$ \\
Buoyancy number            & average, max and min & average & $\times$ \\
Longitudinal buoyancy      & average, max and min & average & average \\
Transverse buoyancy        & average, max and min & average & $\times$ \\
Vanishing artificial diffusion & average, max and min & average & average \\
Transport equation residual  & one value & one value & one value \\
Transport equation residual old & one value & one value & one value \\
Transport equation residual ratio & one value & one value & one value \\
Inner nonlinear iterations & one value & one value & one value \\
\midrule
Number of features:         & 35 & 17 & 14 \\
\bottomrule
\end{tabular}
\end{sc}
\end{small}
\end{center}
\vskip -0.1in
\end{table*}

\subsection{Training dataset}

\begin{figure}[!htb]
	\centering
	\includegraphics[width=1.0\columnwidth]{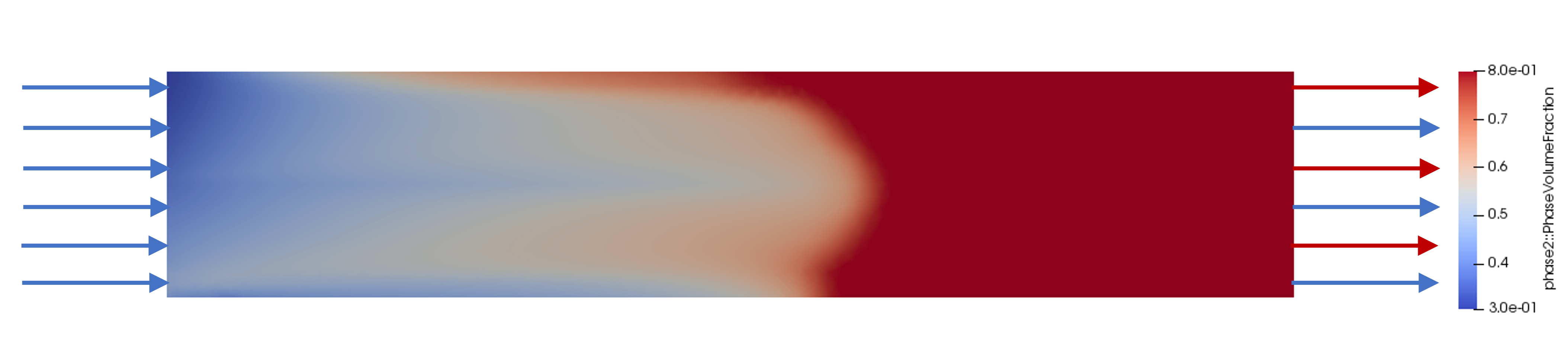}
	\caption{Two-dimensional homogeneous layered reservoir used for offline training. It shows the saturation of the displaced phase in one point in time during the numerical simulation. Blue represents the injected phase and red the displaced phase.}
	\label{fig:simplemodel}
\end{figure}

Offline training is conducted using data from a multiphase flow simulation within a straightforward two-dimensional layered model, as depicted in Figure \ref{fig:simplemodel}. In this setup, one phase is injected from the left side, and both phases are produced on the right side. The objective of this process is to facilitate the injected phase in displacing the existing phase through the porous media.

We run the offline training using data from a multiphase porous media flow simulation of a simple two-dimensional layered model, as shown in Figure \ref{fig:simplemodel}. We inject one phase on the left side and produce both phases on the right side. The aim is for the injected phase to push the displaced phase through the porous media. 
Because the reservoir model must be run numerous times to generate the dataset for training, employing a simple reservoir model considerably simplifies the training process. 

To create a diverse offline training dataset, we perturb the physical parameters listed in Table \ref{tab:inputs} by varying the simulation inputs. These inputs include horizontal and vertical permeabilities, porosity, time step size, capillary pressure, relative permeability, gravity magnitude and direction, viscosity, and the relaxation parameter. We run 6,500 simulations, which represents approximately 1.8 million instances (outer nonlinear iterations). This dataset is split into a training (80\%) and test (20\%) set.

In the online stage (online/adaptive learning), for each outer nonlinear iteration we store: the dimensionless numbers in Table \ref{tab:inputs}, the relaxation factor, and the number of inner nonlinear iterations. After running $W$ outer nonlinear iterations, we then run the online training. We repeat this process every $W$ nonlinear iterations until the end of the numerical simulation.   

\section{Test case description}\label{sec:apptc}

We evaluated the adaptive learning acceleration across four distinct reservoir models as in \citet{silva:21machine}. Each of them represents two-phase immiscible flow within a porous medium:

\begin{enumerate}
    \item Test case 1: two-dimensional layered reservoir model used for training.
    \item Test case 2: heterogeneous two-dimensional reservoir model with distinct permeabilities across its four quadrants.
    \item Test case 3: more intricate three-dimensional reservoir model simulating channelized formations.
    \item Test case 4: complex three-dimensional reservoir model with faulted structures.
\end{enumerate}

\begin{figure*}[!tb]
	\centering
	\begin{subfigure}[t]{0.49\textwidth}
		\centering
		\includegraphics[width=\textwidth]{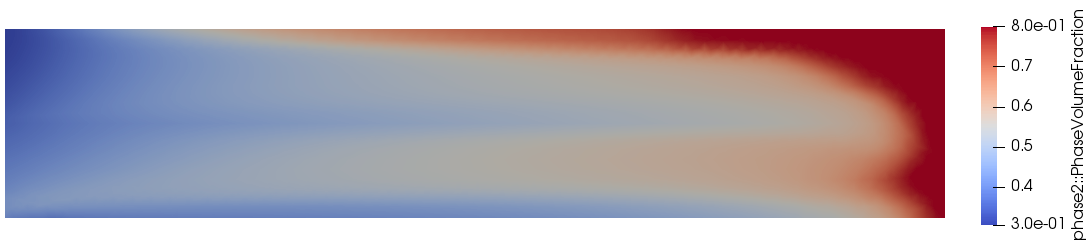}
		\caption{Test case 1 (2D - used for training).}
		\label{fig:tc1}
	\end{subfigure}
	\begin{subfigure}[t]{0.49\textwidth}
		\centering
		\includegraphics[width=\textwidth]{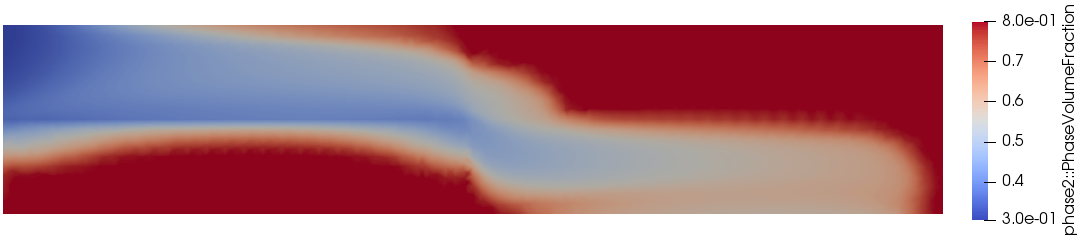}
		\caption{Test case 2 (2D).}
		\label{fig:tc2}
	\end{subfigure}
	\begin{subfigure}[t]{0.49\textwidth}
		\centering
		\includegraphics[width=\textwidth]{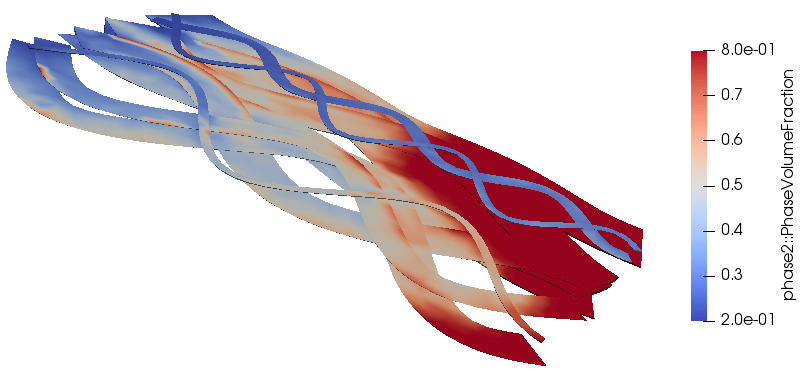}
		\caption{Test case 3 (3D).}
		\label{fig:tc3}
	\end{subfigure}
	\begin{subfigure}[t]{0.49\textwidth}
		\centering
		\includegraphics[width=\textwidth]{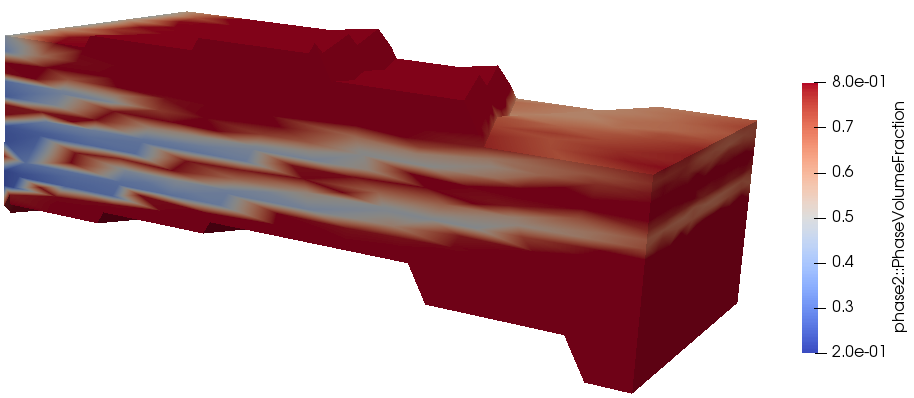}
		\caption{Test case 4 (3D).}
		\label{fig:tc4}
	\end{subfigure}
	\caption{Saturation of the displaced phase in one point in time during the numerical simulation. Blue represents the injected phase and red the displaced phase. In all cases, we injected one phase on the left and produce both phases on the right.}
	\label{fig:testcases}
\end{figure*}

For all reservoir models, both gravity and capillary pressure were considered. The capillary pressure and relative permeability were modeled using the Brooks–Corey parametrization \cite{brooks:64} with the values shown in Table \ref{mla_tab:rockfluid}. In our tests, we injected one phase from the left and extracted both phases from the right. Figure \ref{fig:testcases} provides a saturation map for each test case.

\begin{table}[!h] 
\caption{Rock-fluid modelling using the Brooks–Corey parametrization.}
\label{mla_tab:rockfluid}
    \vskip 0.15in
    \begin{center}
    \begin{small}
    \begin{sc}
    \begin{tabular}{lcccc}
    \toprule
		\textbf{Property} & \textbf{Cases 1 and 2}  &  \textbf{Case 3} &  \textbf{Case 4} \\
	\midrule
		Relative permeability end point of water & 1.0 & 0.3 & 0.3  \\
		Relative permeability end point of oil & 1.0 & 0.8 & 0.8  \\
		Relative permeability exponent & 2.0 & 2.0 & 2.0  \\
		Immobile fraction of water & 20 \% & 20 \% & 20 \% \\
		Immobile fraction of oil & 30 \% & 20 \% & 20 \% \\
		Entry capillary pressure & 1000 \text{Pa} & 100 \text{Pa} & 10000 \text{Pa} \\			
		Capillary exponent & 1.0 & 1.0 & 1.0\\
    \bottomrule
    \end{tabular}
    \end{sc}
    \end{small}
    \end{center}
    \vskip -0.1in
\end{table}


The convergence criteria for the nonlinear solver are defined as follows: the relative mass conservation of the system must be less than $10^{-3}$, and the infinite norm of the delta saturation between two successive nonlinear iterations should be less than $10^{-2}$ within a single time step. Additionally, we set maximum limits of 30 and 10 for the outer and inner nonlinear iterations, respectively. In evaluating different scenarios, the total number of nonlinear iterations is determined by summing the outer iterations and one-third of the inner iterations. This approach aligns with the methodologies previously established by \citet{salinas:17} and \citet{silva:21machine}.


\subsection{Test case 1}

\begin{figure}[!tb]
	\centering
	\includegraphics[width=0.8\columnwidth]{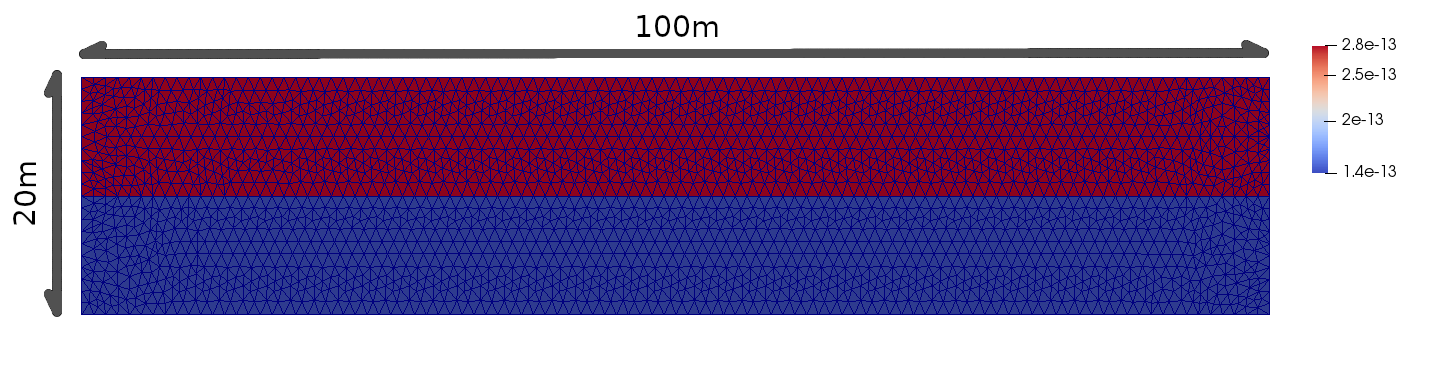}
	\caption{Test case 1 (2D) permeability field in m$^2$.}
	\label{fig:c1model}
\end{figure}

Test case 1 is the layered reservoir model used to generate the dataset for the offline training. The model comprises two layers and is built in two-dimensions (Figure \ref{fig:c1model}). 
The upper layer displays a porosity of 10\% and a horizontal permeability measuring 200 millidarcies (mD). Conversely, the lower layer possesses a porosity of 20\% and a horizontal permeability of 100 mD. The vertical permeability in both layers is 10\% of their respective horizontal permeabilities. Additionally, the viscosity ratio of the injected fluid to the displaced fluid is 5 and the density contrast between these two fluids is 300 $kg/m^3$.


\subsection{Test case 2}

\begin{figure}[!tb]
	\centering
	\includegraphics[width=0.8\columnwidth]{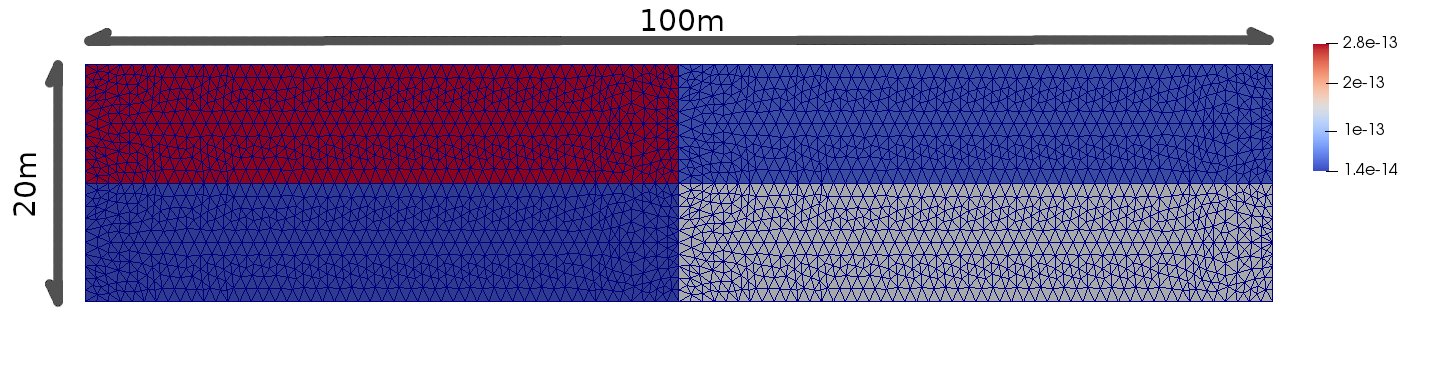}\\
	\caption{Test case 2 (2D) permeability field in m$^2$.}
	\label{fig:c2model}
\end{figure}

Test case 2 is a heterogeneous two-dimensional reservoir model with distinct permeabilities across its four quadrants (Figure \ref{fig:c2model}). 
In the upper layer, the horizontal permeabilities are 200 mD and 20 mD, from left to right. In contrast, the lower layer exhibits horizontal permeabilities of 10 mD and 100 mD. Across both layers, the vertical permeability equals the horizontal permeability.
Additionally, a uniform porosity of 20\% is observed across all layers. Regarding fluid properties, a viscosity ratio of 5 exists between the displaced and the injected fluid, and there is a density contrast of 300 kg/m³ between fluids. 

\subsection{Test case 3}

\begin{figure}[!tb]
	\centering
	\includegraphics[width=0.78\columnwidth]{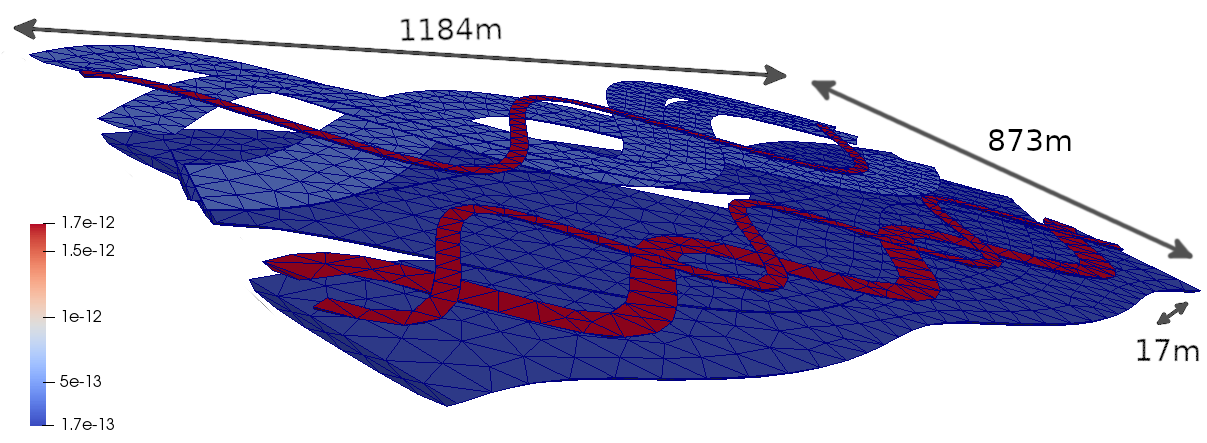}
	\caption{Test case 3 (3D) permeability field in m$^2$.}
	\label{fig:c3model}
\end{figure}

Test case 3 is a three-dimensional reservoir model featuring fluvial sandstone channels situated within a backdrop of low-permeability mudstone (Figure \ref{fig:c3model}). The channels are classified into three types based on their width: thin, medium, and wide. The permeabilities assigned to these types are 1000 mD for thin channels, 200 mD for medium-sized channels, and 100 mD for wide channels. In each channel type, the permeability in the vertical direction matches the horizontal permeability. The porosity for all channels is uniformly set at 20\%.
Further parameters include a viscosity ratio of 5 between the displaced and injected fluid, and a density contrast of 289 kg/m³ between fluids. 


\subsection{Test case 4}

\begin{figure}[!tb]
	\centering
	\includegraphics[width=0.75\columnwidth]{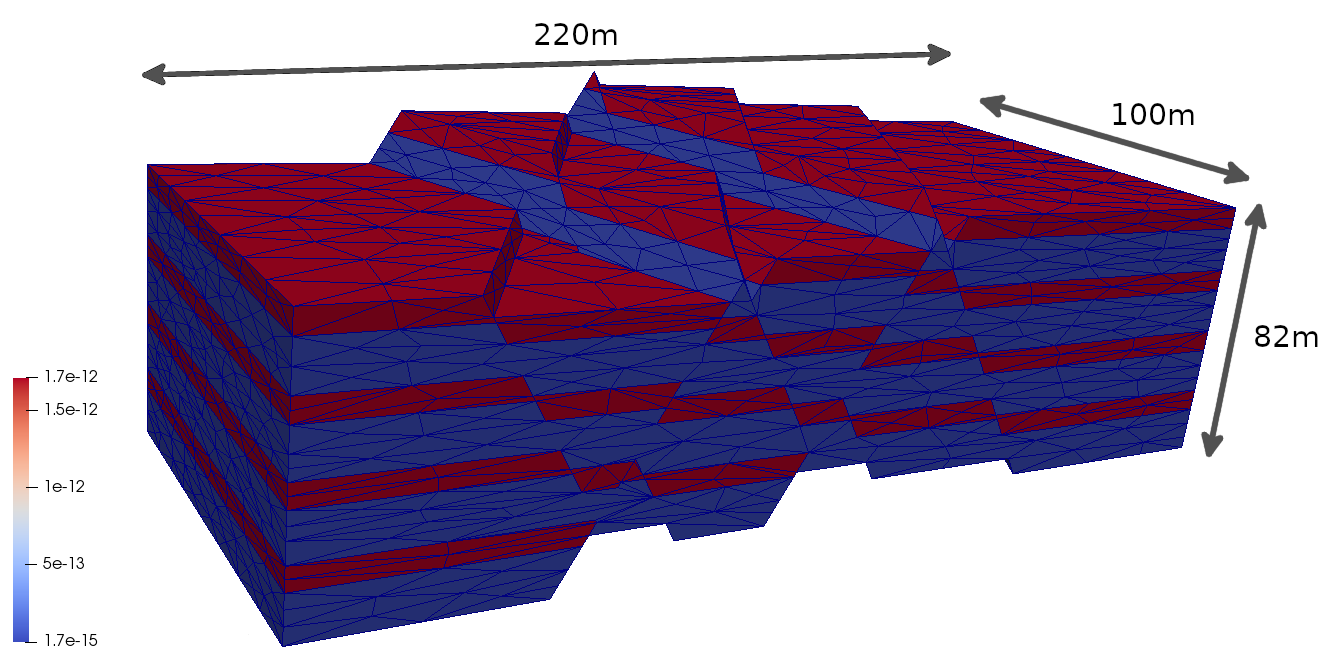}
	\caption{Test case 4 (3D) permeability field in m$^2$.}
	\label{fig:c4model}
\end{figure}

Test case 4 features a three-dimensional, complex reservoir model characterized by faulted geological structures (Figure \ref{fig:c4model}). 
The reservoir is characterized by a sequence of alternating layers of mudstone and sandstone. The mudstone layers have a horizontal permeability of 1 mD, while the sandstone layers have a horizontal permeability of 1000 mD. In both layer types, the permeability in the vertical direction matches the horizontal permeability. Regarding porosity, the mudstone layers exhibit a porosity of 20\%, while the sandstone layers display a lower porosity of 10\%. Moreover, a viscosity ratio of 5 is present between the displaced and the injected fluid accompanied by a phase density contrast of 300 kg/m³. 

\section{Numerical results}\label{sec:exp}


In the following sections, we test and select the best set of dimensionless numbers as input parameters, and analyse several machine learning models in terms of accuracy, prediction time and simulation results. Following that, we apply and analyse the proposed online/adaptive learning acceleration. Finally, we integrate the best machine learning model into IC-FERST in order to actually generate a walltime reduction. As a baseline for comparison, we use here the nonlinear solver acceleration presented in our previous work \citep{silva:21machine}.There we used a random forest to calculate the relaxation parameter, but relying only on the offline training. For the remainder of this paper, the random forest model used in \citet{silva:21machine} will be referred as original random forest. 

\subsection{Machine learning feature selection}\label{sec:feature}

We investigate different sets of features, in order to evaluate their effect on the proposed method. In \citet{silva:21machine} a set of 35 features were chosen (see Table~\ref{tab:inputs}), and for the parameters defined in each grid cell the average, minimum and maximum over the domain were calculated, excepted for the Courant–Friedrichs–Lewy (CFL) number where it is already known that the maximum value plays an important role in the nonlinear solver convergence \citep{younis2010adaptively,salinas:17,Obeysekara2021}. In this work, we test two new sets of features as shown in Table \ref{tab:inputs}. We notice (from the dataset) that the average, minimum and maximum values are highly correlated when considering the same parameter. Therefore, in the first set (Set 1) we consider only the average. For the second set (Set 2), we have looked the feature importance of the original random forest, and for the capillary and buoyancy numbers we included only the most important features. As for the original random forest, we run a hyperparameter optimization using a grid search with a 3-fold cross-validation for the two new sets of features. The implementation of the random forests used here is the one in \citet{scikit-learn}. For the sake of comparison, the results reported in this section only consider the offline training.     

\begin{figure}[!tb]
	\centering
	\includegraphics[width=0.9\columnwidth]{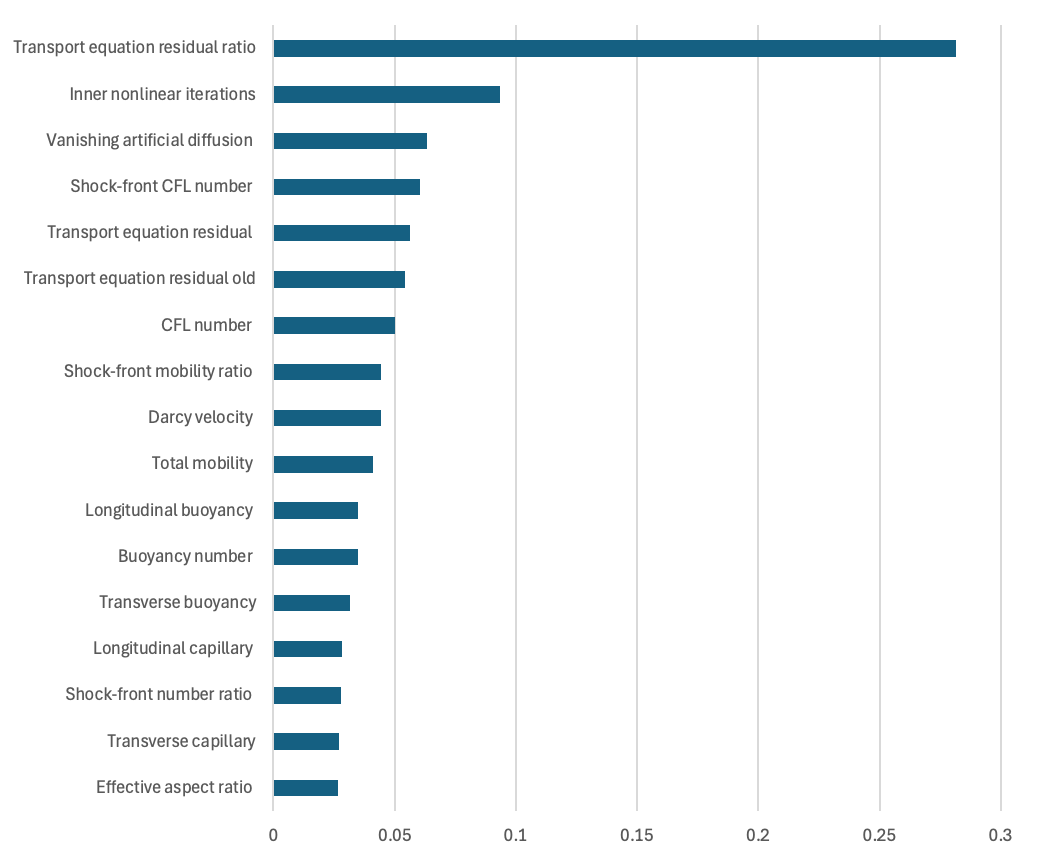}
	\caption{Feature importance of SET 1.}
	\label{fig:featimp}
\end{figure}

In Figure~\ref{fig:featimp}, we show the feature importance of the SET 1 generated by the original random forest. We can notice that the transport equation residual ratio (how fast the solver is converging) is the most important feature, while the other residuals also appear among the top variables. Moreover, we see that the Shock-front CFL number and CFL number appear among the most important features. This CFL dependence on the solver convergence is also seen in \citep{younis2010adaptively,salinas:17,Obeysekara2021}. Figure~\ref{fig:featimp} also shows that the most important features among the buoyancy and capillary numbers are the longitudinal buoyancy and longitudinal capillary. Because of that, we kept them on SET 2, as mentioned in the previous paragraph.

\begin{figure}[!tb]
	\centering
	\begin{subfigure}[t]{0.45\textwidth}
		\centering
		\includegraphics[width=\textwidth]{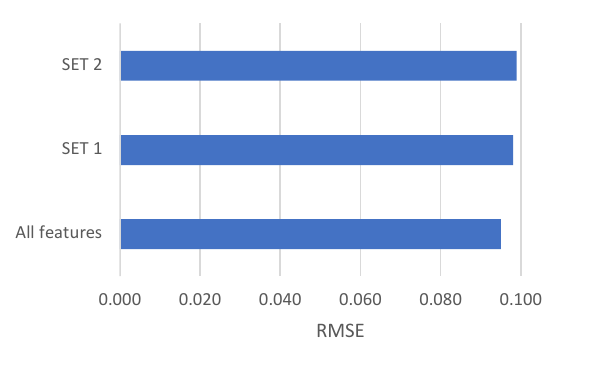}
		\caption{}
		\label{fig:frmse}
	\end{subfigure}
	\begin{subfigure}[t]{0.45\textwidth}
		\centering
		\includegraphics[width=\textwidth]{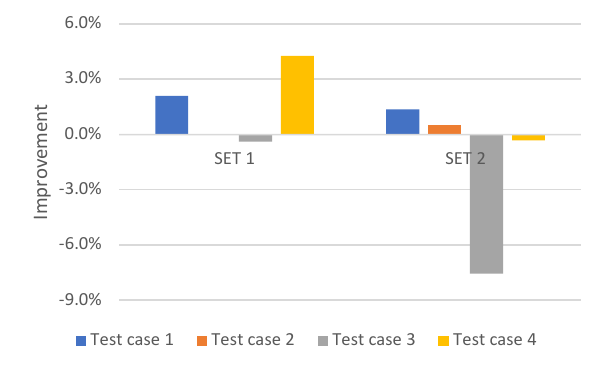}
		\caption{}
		\label{fig:fcases}
	\end{subfigure}
	\begin{subfigure}[t]{0.45\textwidth}
		\centering
		\includegraphics[width=\textwidth]{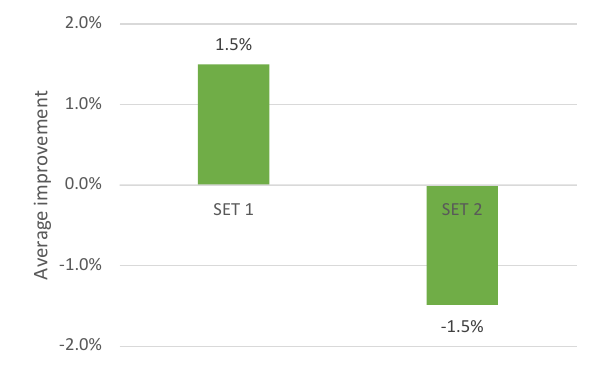}
		\caption{}
		\label{fig:faverage}
	\end{subfigure}
	\caption{Comparison of different number of features. (a) the test root mean square error of the trained random forest models. (b) the improvement in the number on nonlinear iterations compared with the original random forest. (c) average improvement (reduction) in the number of nonlinear iterations compared with the original random forest.}
	\label{fig:features}
\end{figure}

Figure \ref{fig:features} shows a comparison of the different number of features. In Figure \ref{fig:frmse} we can see the test root mean square error (RMSE) of the random forest models for the three sets of features. There is no significant difference in the RMSE when comparing the models. However, in Figures \ref{fig:fcases} and \ref{fig:faverage} we can notice that the reduction in the number of nonlinear iterations (improvement) is greater for Set~1 (fewer features). Set~2 (many fewer features) increased the average number of nonlinear iterations. 
The results show that the acceleration using the random forest with Set~1 performed better than the original random forest and the one with Set~2. Therefore, the 17 features of Set 1 were selected to continue the experiments. 

\subsection{Machine Learning model selection}

\begin{figure}[!tb]
	\centering
	\begin{subfigure}[t]{0.49\textwidth}
		\centering
		\includegraphics[width=\textwidth]{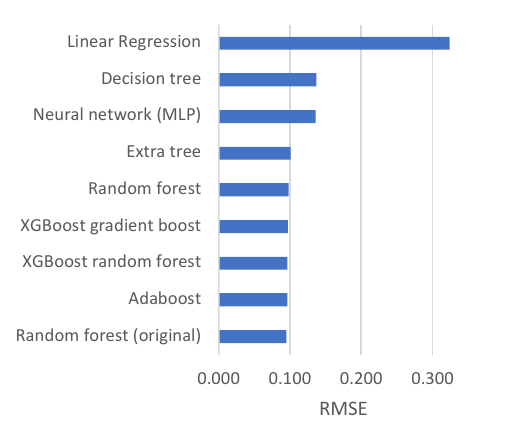}
		\caption{}
		\label{fig:model1}
	\end{subfigure}
	\begin{subfigure}[t]{0.49\textwidth}
		\centering
		\includegraphics[width=\textwidth]{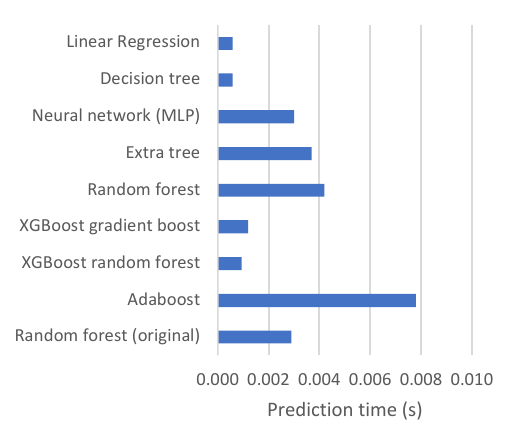}
		\caption{}
		\label{fig:model2}
	\end{subfigure}
	\caption{Comparison of different machine learning models. (a) the test root mean square error. (b) the one-instance prediction time}
	\label{fig:model}
\end{figure}

We compare several machine learning models for controlling the numerical relaxation. All of them were implemented using \citet{scikit-learn}, except for the xgboost models where we used \citet{xgboost} and the neural network (multilayer perceptron - MLP) where we use \citet{tensorflow:2015}. The input in all the models are the 17 features selected in the previous section (Set 1), except for the original random forest that uses the 38 features in Table \ref{tab:inputs}. For each machine learning model we run a hyperparameter optimization 
using a grid search with a 3-fold cross-validation. Again, for the sake of comparison, the results reported in this section only consider the offline training. 
Figure \ref{fig:model} shows a comparison of the different models. In Figure \ref{fig:model1} we can see the test RMSE of the machine learning models. Apart from the linear regression 
all the models present similar RMSE. Figure \ref{fig:model2} shows the prediction time for one instance. For each machine learning model, we run the prediction 1000 times and average the resulted prediction time. As we use the machine learning model to generate the relaxation parameter in each outer nonlinear iteration, we want the prediction time to be as low as possible. We can notice that the xgboost models, linear regression and decision tree exhibit the lowest prediction times. 

\begin{figure}[!tb]
	\centering
	\begin{subfigure}[t]{0.99\textwidth}
		\centering
		\includegraphics[width=\textwidth]{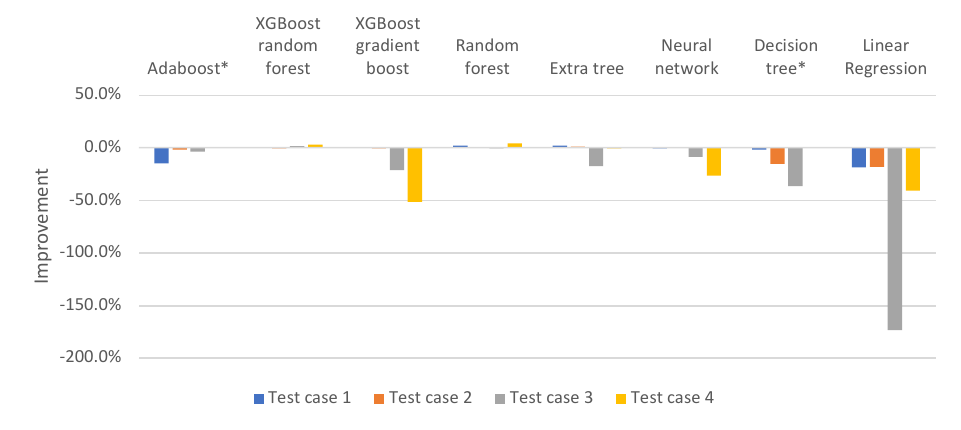}
		\caption{}
		\label{fig:impmodel1}
	\end{subfigure}
	\begin{subfigure}[t]{0.99\textwidth}
		\centering
		\includegraphics[width=\textwidth]{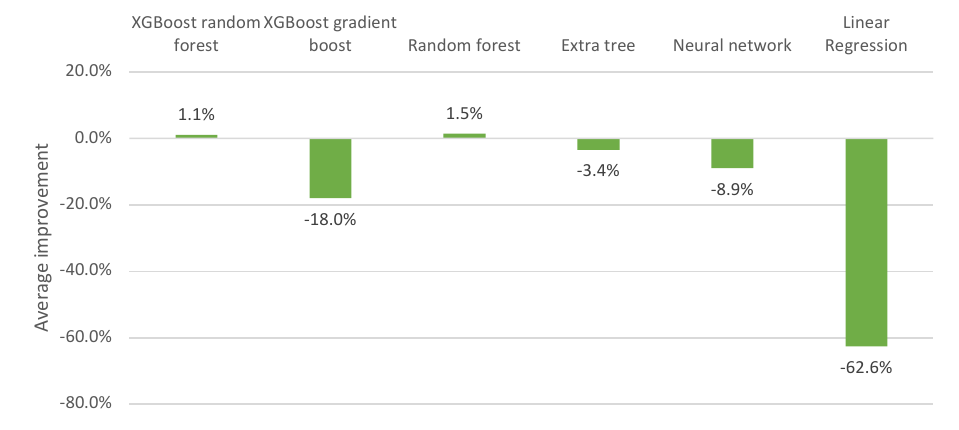}
		\caption{}
		\label{fig:impmodel2}
	\end{subfigure}
	\caption{Comparison of different machine learning models. (a)  Improvement in the number on nonlinear iterations compared with the original random forest. (b) Average improvement (reduction) in the number of nonlinear iterations compared with the original random forest.}
	\label{fig:impmodel}
\end{figure}

\begin{figure*}[!tb] 
	\centering
	\begin{subfigure}[t]{0.49\textwidth}
		\centering
		\includegraphics[width=\textwidth]{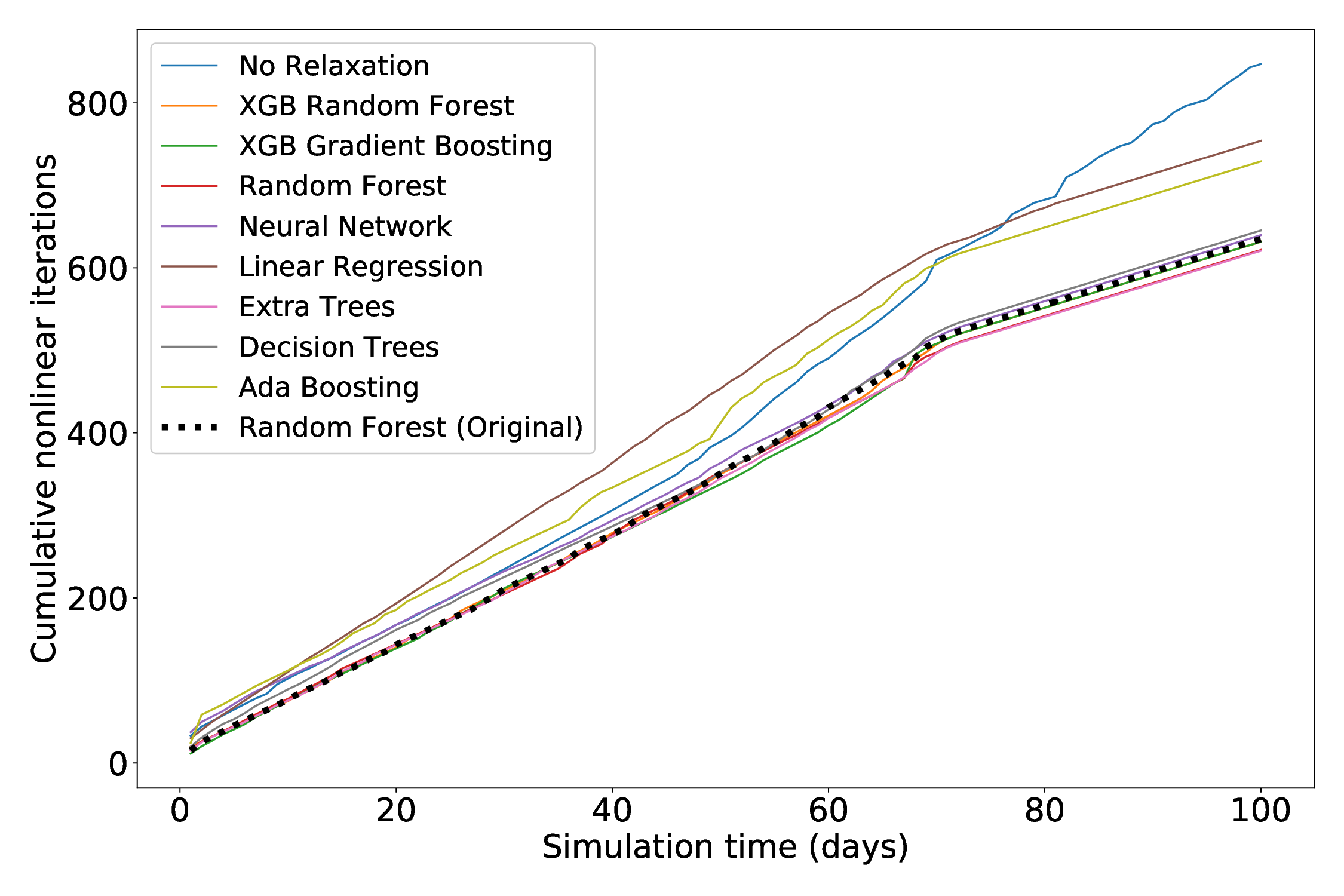}
		\caption{Test case 1.}
		\label{fig:c21}
	\end{subfigure}
	\begin{subfigure}[t]{0.49\textwidth}
		\centering
		\includegraphics[width=\textwidth]{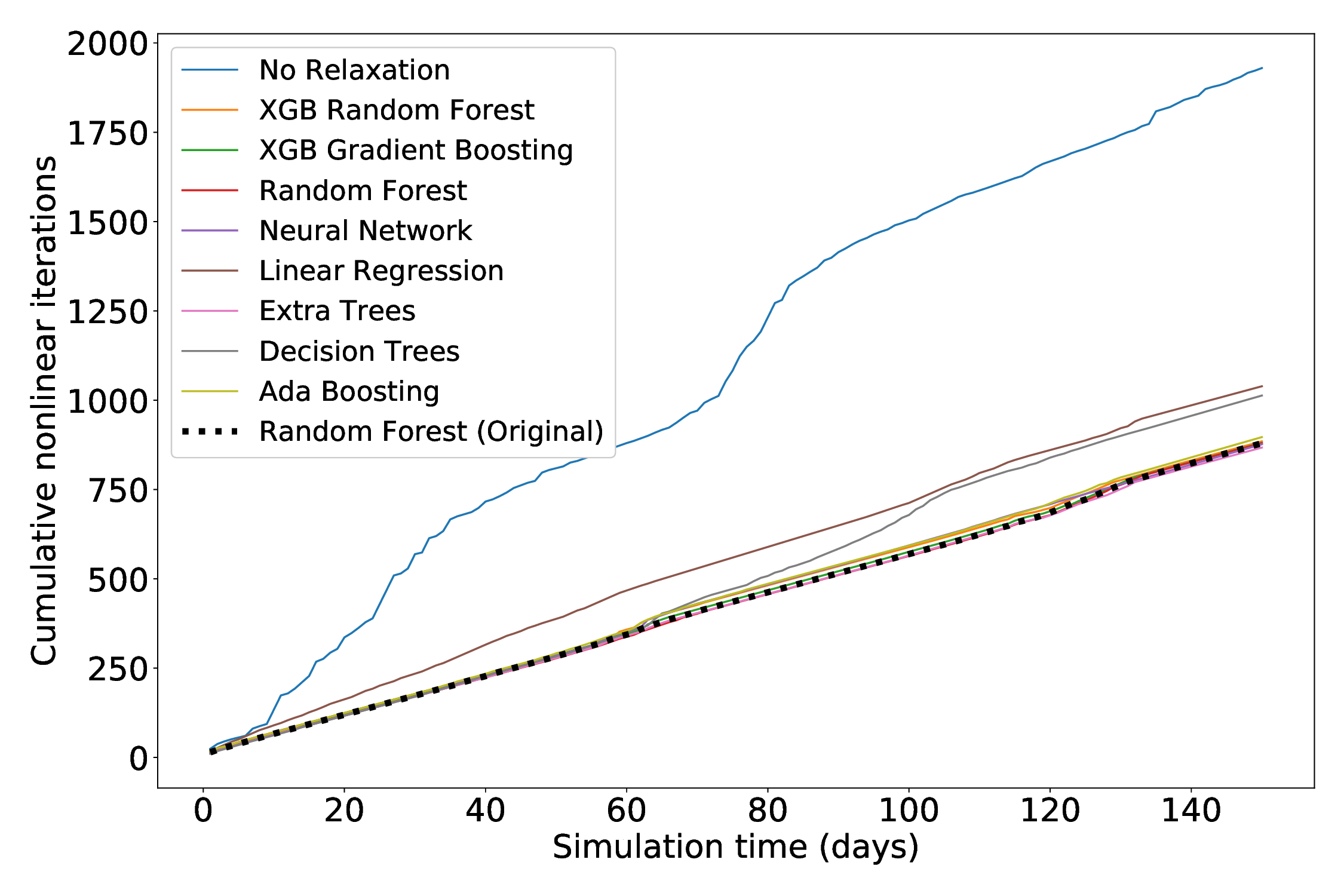}
		\caption{Test case 2.}
		\label{fig:c22}
	\end{subfigure}
	\begin{subfigure}[t]{0.49\textwidth}
		\centering
		\includegraphics[width=\textwidth]{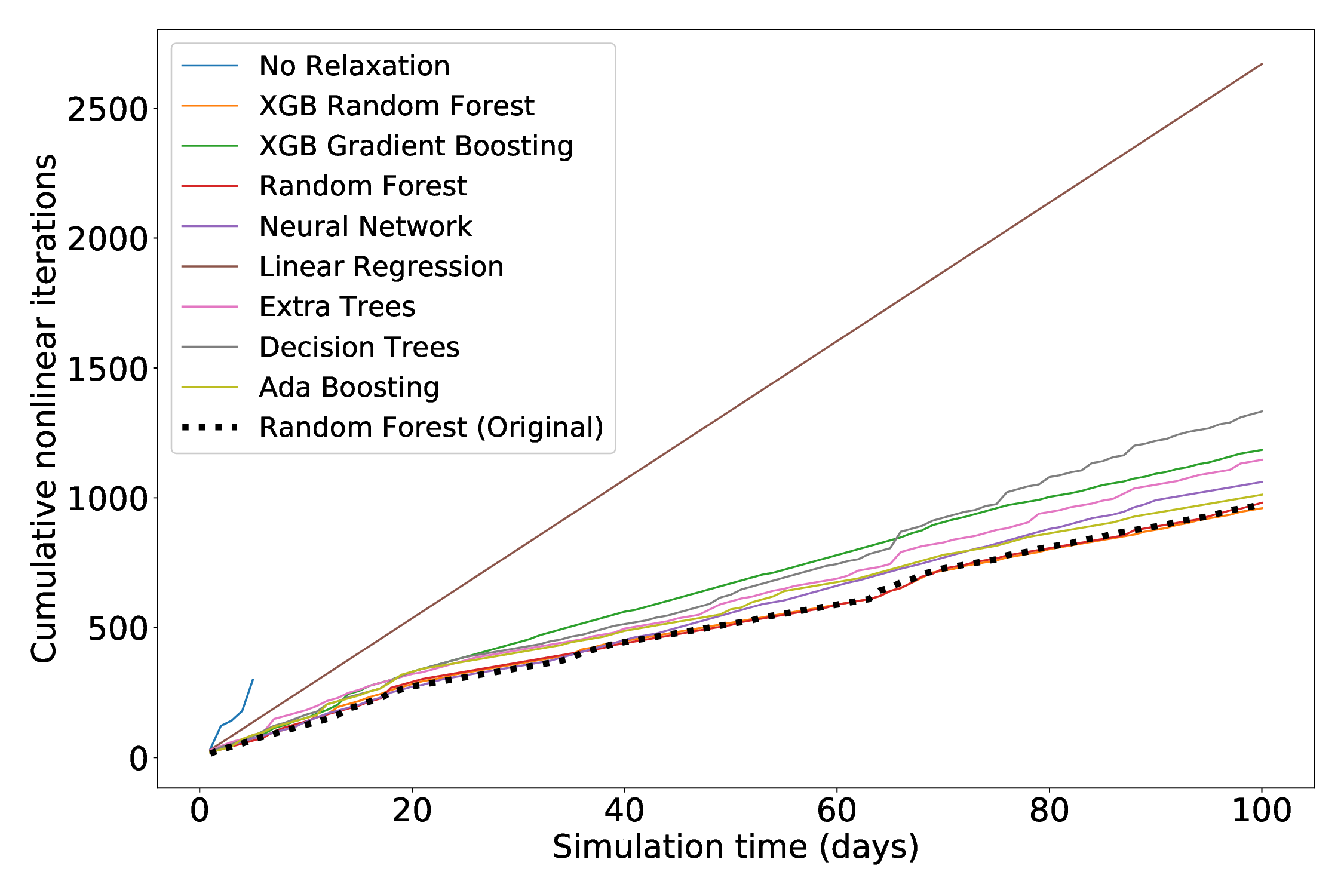}
		\caption{Test case 3.}
		\label{fig:c23}
	\end{subfigure}
	\begin{subfigure}[t]{0.49\textwidth}
		\centering
		\includegraphics[width=\textwidth]{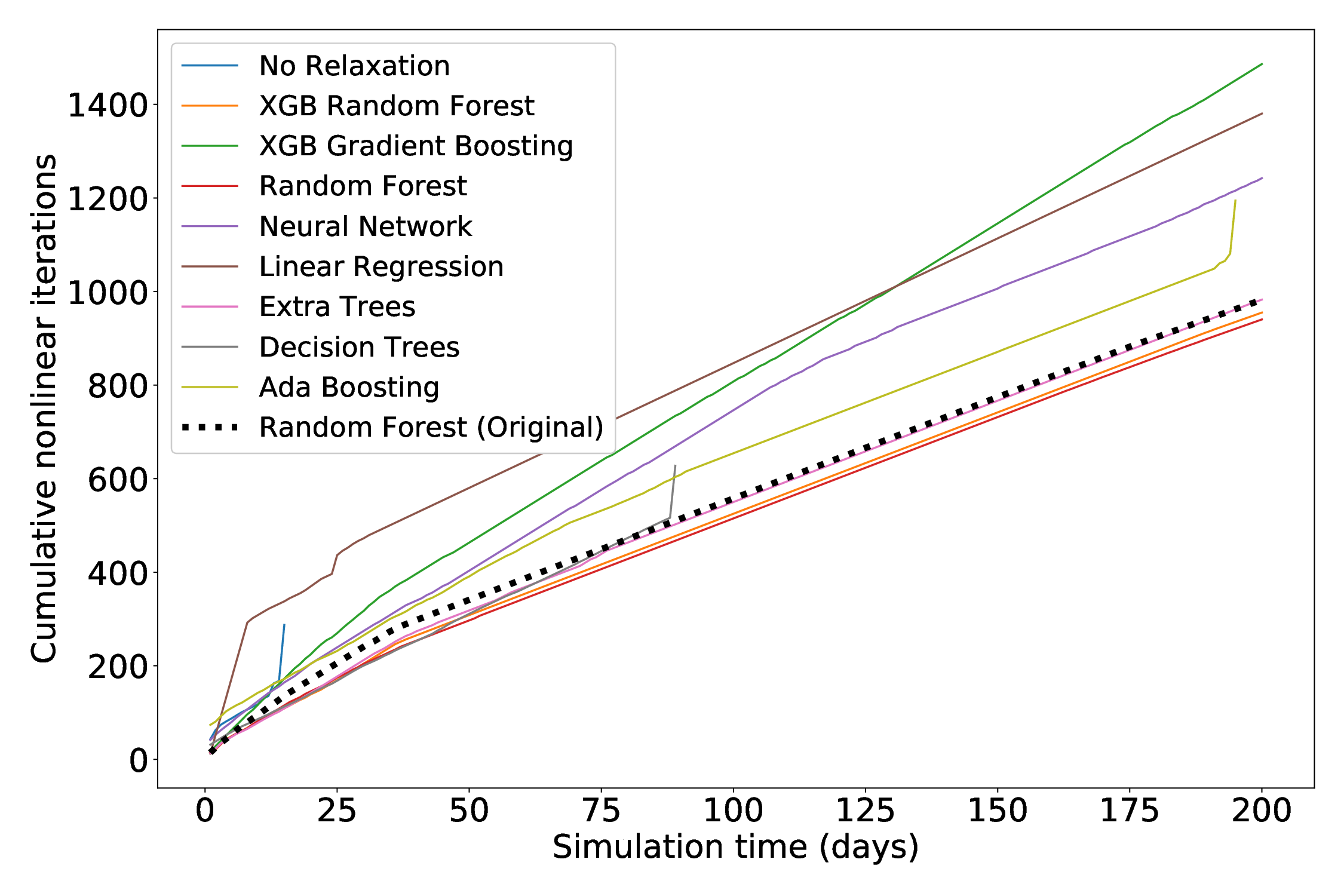}
		\caption{Test case 4.}
		\label{fig:c24}
	\end{subfigure}
	\caption{Cumulative number of nonlinear iterations for different machine learning models and for the case with no relaxation.}
	\label{fig:cummodel}
\end{figure*}

Figure \ref{fig:impmodel} shows the reduction in the number of nonlinear iterations (improvement) for all the machine learning models, using as baseline the original random forest. We can also see the cumulative number of nonlinear iterations during the numerical simulation for all the machine learning models and the case with no relaxation in Figure \ref{fig:cummodel}. In Figure \ref{fig:impmodel1} the solutions accelerated using the adaboost and decision tree have failed to converge for test case 4. Because of that, their average improvement is not shown in Figure \ref{fig:impmodel2}. The only models that have generated better results than the original random forest are the xgboost random forest and the random forest (both with Set 1). It is also worth noting (from Figure~\ref{fig:cummodel}) that the approach with no relaxation generated the worst results for the test cases 1 and 2, and it failed to converge for the test cases 3 and 4. 

\subsection{Online learning}\label{sec:online} 



In the previous sections, we focused on the offline training. The two machine learning models that generated better results than the original random forest are the xgboost random forest and the random forest. Adding that ensemble learners have the advantage of being flexible, as new learners can be selectively added or removed, that they are usually easier to optimize, and that they are often used when learning from evolving data streams \citep{oza2001online,read2012batch,gomes2017adaptive,montiel2020adaptive}, we have selected these two models to test the online/adaptive learning shown in Figure \ref{fig:mlacc1}. 

Many state-of-the-art ensemble methods for data stream learning are adapted versions of bagging and boosting strategies \citep{gomes2017adaptive,montiel2020adaptive}. In this work, given that the XGBoost library \citep{xgboost} already supports a boosting strategy for training continuation, we apply the online learning method Adaptive XGBoost, proposed by \citet{montiel2020adaptive}, to the xgboost model. Nonetheless, we consider here that we already have the first learner trained offline. In the online stage, as new outer nonlinear iterations arrive, they are stored in a buffer of size $W$. Once the buffer is full, we train a new member of the ensemble with the residuals from the previous members. It is worth noting that we do not reach the maximum ensemble size, unlike \citet{montiel2020adaptive}, because the time of the numerical simulation is finite the same way as the number of updates (usually less than 10 updates by numerical simulation). For the random forest, we modify the method proposed by \citet{montiel2020adaptive} to work as a bagging strategy, since this is the strategy supported by the random forest in the Scikit-learn library \citep{scikit-learn}. As in the Adaptive XGBoost method, as new outer nonlinear iterations arrives, they are stored in a buffer of size $W$. The difference in the random forest is that once the buffer is full, we train a new set of ensemble members with the samples in the buffer, and add it to the previous ensemble. The hyperparameters used to configure this methods are described in Appendix~\ref{sec:apphyper}.  

\begin{figure}[!tb]
	\centering
	\includegraphics[width=0.7\columnwidth]{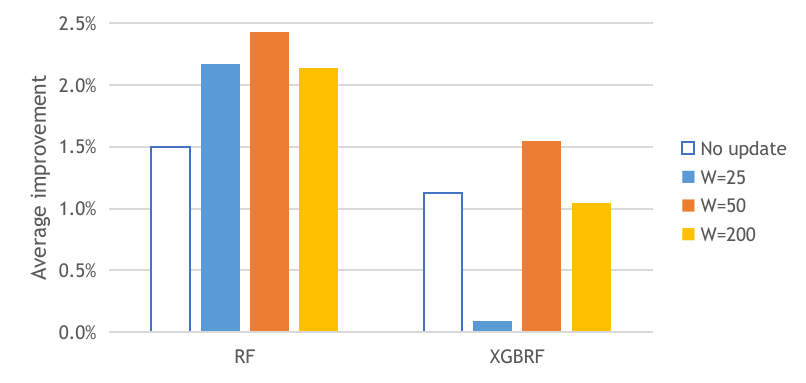}
	\caption{Average improvement (reduction) in the number of nonlinear iterations compared with the original random forest. The Base case is considering no online updating.}
	\label{fig:alc_avg}
\end{figure}

We applied these two strategies to the four test cases shown in Figure \ref{fig:testcases}. Tables \ref{tab:rf} and \ref{tab:xgbrf} show the percentage reduction (improvement) in the number of nonlinear iterations compared with the original random forest (state-of-the-art results from \citet{silva:21machine}). The columns RF and XGBRF represent the two machine learning models selected from the previous section, but without the online learning stage. In the batch-incremental learning, the size of the batch must be chosen to provide a balance between accuracy (large batches) and response to new instances (smaller batches) \citep{read2012batch}. In this context, we test different buffer sizes ($W$) represented in the remaining columns of Tables \ref{tab:rf} and \ref{tab:xgbrf}. Figure \ref{fig:alc_avg} also shows the average reduction of each one of these strategies. We also notice that the strategy with buffer size equals 50 ($W=50$) produced the best average performance for both machine learning models. 
Also for the strategy with buffer size equals 25 ($W=25$), the number of nonlinear iterations became worse than the original random forest for some test cases. This can indicate that updating the machine learning model with short buffer sizes does not necessarily improve the final results. 

\begin{table}[!tb]
\caption{Adaptive learning acceleration using the random forest, and a bagging strategy for the online learning. Improvement (reduction) in the number on nonlinear iterations compared with
the original random forest.}
\label{tab:rf}
\vskip 0.15in
\begin{center}
\begin{small}
\begin{sc}
\begin{tabular}{lcccc}
\toprule
TEST CASE & RF & $W$=25 & $W$=50 & $W$=200  \\
\midrule
CASE 1 & 2.1\% & 2.8\% & 3.7\%  & 1.2\% \\
CASE 2 & 0.0\% & -1.8\% & 0.6\% & 0.5\% \\
CASE 3 & -0.4\% & 2.7\% & 0.9\% & 2.7\% \\
CASE 4 & 4.3\% & 5.0\% & 4.5\%  & 4.2\% \\
\midrule
Avg. Reduction  & 1.5\% & 2.2\% & 2.4\% & 2.1\% \\
\bottomrule
\end{tabular}
\end{sc}
\end{small}
\end{center}
\vskip -0.1in
\end{table}

\begin{table}[!tb]
\caption{Adaptive learning acceleration using the xgboost random forest, and a boosting strategy for the online learning. Improvement (reduction) in the number on nonlinear iterations compared with
the original random forest.}
\label{tab:xgbrf}
\vskip 0.15in
\begin{center}
\begin{small}
\begin{sc}
\begin{tabular}{lcccc}
\toprule
TEST CASE & XGBRF  & $W$=25 & $W$=50 & $W$=200   \\
\midrule
CASE 1 & 0.5\% & -0.4\% & 2.4\% & 0.8\% \\
CASE 2 & -0.6\% & -1.3\% & 0.0\% & 0.0\% \\
CASE 3 & 1.8\% & -1.3\% & 0.4\% & 0.4\% \\
CASE 4 & 2.8\% & 3.4\% & 3.3\% & 2.9\% \\
\midrule
Avg. Reduction  & 1.1\% & 0.1\% & 1.5\% & 1.0\% \\
\bottomrule
\end{tabular}
\end{sc}
\end{small}
\end{center}
\vskip -0.1in
\end{table}

\begin{figure}[!tb]
	\centering
	\includegraphics[width=\columnwidth]{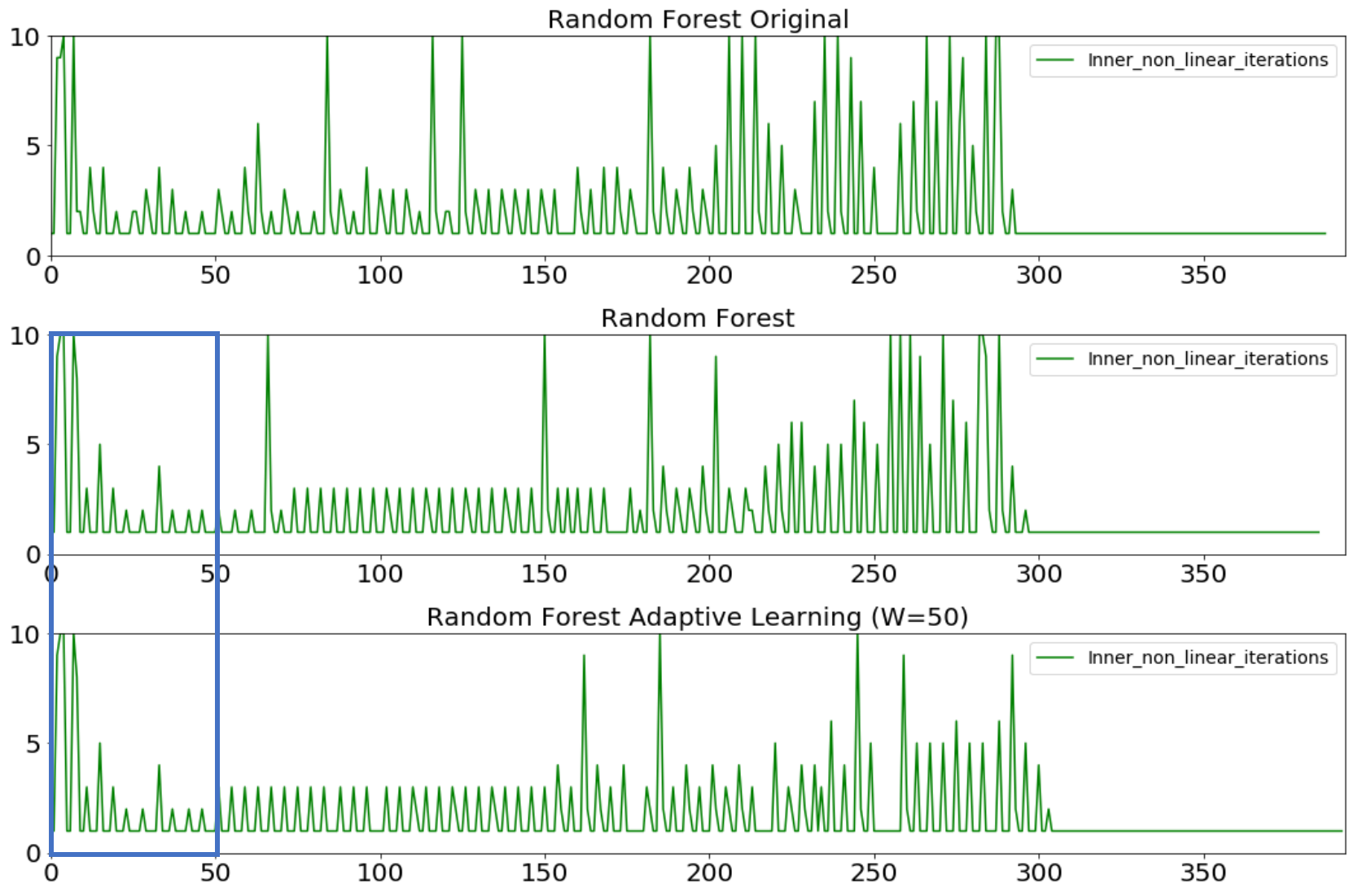}
	\caption{Comparison of the number of inner nonlinear iteration over the simulation period, for test case 1. The horizontal axes are the outer nonlinear iterations and the vertical axes the number of inner nonlinear iterations.}
	\label{fig:nli}
\end{figure}

\begin{figure}[!tb]
	\centering
	\includegraphics[width=\columnwidth]{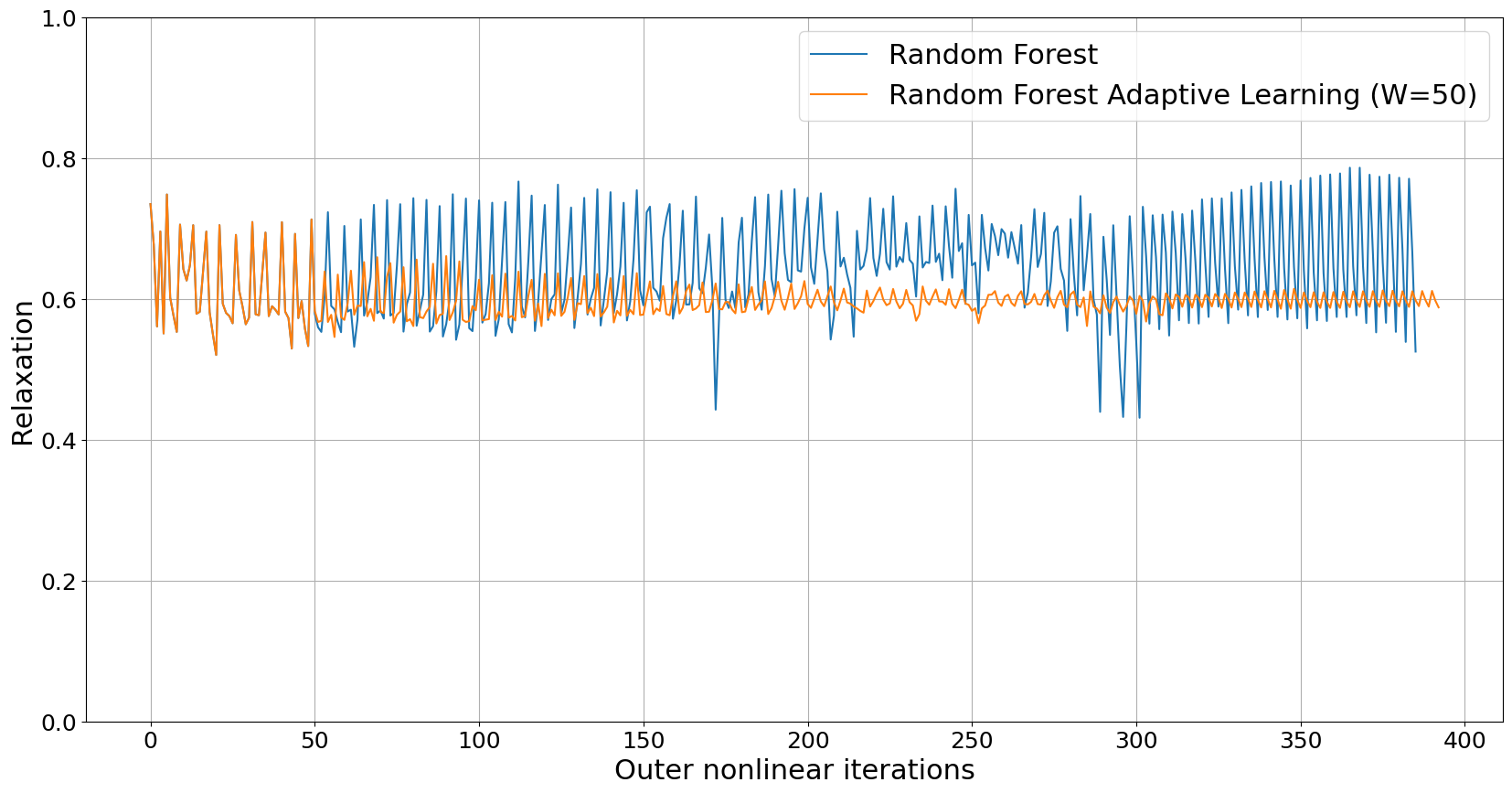}
	\caption{Values of the relaxation parameter over the simulation period, for test case 1.}
	\label{fig:rel}
\end{figure}

Figure \ref{fig:nli} shows the comparison of the number of inner nonlinear iterations over the simulation period, for the test case~1. 
The horizontal axes represent the outer nonlinear iterations. We can notice from the blue box on the left, that for the first 50 outer nonlinear iterations the results from the random forest with the online/adaptive learning and the one without it are the same. This is because the buffer has size 50 ($W=50$), then until it is full no update is performed in the machine learning model. We can also see a clear reduction in the inner nonlinear iterations from the original random forest to the online/adaptive learning. It is worth noting that approximately after 300 outer nonlinear iterations the shock front (the interface between the two phases in the porous media, see Figure~\ref{fig:simplemodel}) has passed the simulation domain, making the nonlinear solver convergence much easier for the three strategies, only one inner nonlinear iteration is required to reach the convergence criteria. In Figure~\ref{fig:rel}, we show the values of the relaxation parameter over the outer nonlinear iterations for the random forest with and without online/adaptive learning. It shows that the case with the online/adaptive learning reduces the high-frequency variations in the relaxation. During the online learning, the sample training space comprises only the previous iterations, this maybe guiding the machine learning model to choose the best relaxation within this range.

\subsection{Direct coupling} 

All of the previous machine learning models, including the original random forest \citep{silva:21machine}, were not properly coupled with the numerical simulator in Fortran. An external Python script was called in each outer nonlinear iteration to generate the relaxation factor and it was passed ``on the fly'' to the numerical simulator. For that reason, no actual reduction in walltime was observed. Analysing the prediction time (Figure \ref{fig:model2}), the reduction in the number of nonlinear iterations (Figure \ref{fig:impmodel}), and considering the viability to integrate the machine learning code with the numerical simulator (IC-FERST) in Fortran, we chose the xgboost random forest to perform the coupling. A Fortran API for the xgboost was developed (available at \url{https://github.com/viluiz/solver/tree/main/xgboost_coupling}), and we were able to load the machine learning model in memory and call it in each outer nonlinear iteration using a Fortran code. Although, the online training of the machine learning model was not yet supported in the API, thus the results in this section consider $W=\infty$, which means no online update (the online update would further improve the results). Table \ref{tab:runtime} shows the reduction in the number of nonlinear iterations and walltime, when we compare the proposed method with the default dynamic relaxation presented in IC-FERST \citep{salinas:17}. It is a dynamic relaxation that adjusts the relaxation parameter mainly based on the CFL number. It has demonstrated great improvements in the convergence rate of the solver for a wide range of cases \citep{salinas:17}, and up to this point was the best available approach to calculate the relaxation in IC-FERST. The method proposed here is able to outperform this approach, showing actual run time improvements. The results in table \ref{tab:runtime} show that the reduction in walltime is, in some cases, even greater than the reduction in the number of nonlinear iterations. It means that for some cases, the proposed method not only reduces the number of nonlinear iterations, but also makes faster the solution of each nonlinear iteration.           

\begin{table}
\caption{Reduction in run time and number of nonlinear iterations. Comparison between the proposed method (with no online update) and the default dynamic relaxation presented in IC-FERST \citep{salinas:17}.}
\label{tab:runtime}
\vskip 0.15in
\begin{center}
\begin{sc}
\begin{tabular}{lccr}
\toprule
Case & Run time & Nonlinear iterations \\
\midrule
Test case 1    & 19\% & 13\% \\
Test case 2    & 34\% & 37\% \\
Test case 3    & 9\% & 10\% \\
Test case 4    & 85\% & 81\% \\
\bottomrule
\end{tabular}
\end{sc}
\end{center}
\vskip -0.1in
\end{table}

\section{Discussion}\label{sec:disc}

Dealing with nonlinearities in systems of partial differential equations is always challenging and time-consuming, with a high impact on the overall applicability and performance of the nonlinear solver. 
Developing an efficient and robust nonlinear solver is nontrivial and requires a good understanding of the underlying physics of the system being modelled.
The presented technique shows a promising route towards creating a robust and efficient nonlinear solver easily by using a machine learning approach in combination with a numerical relaxation. The approach is simple to develop and quick to implement as it requires only three steps:
\begin{itemize}
    \item Identify dimensionless numbers and the parameter space to study.
    \item Train a machine learning model offline using a simple numerical model to generate the training set.
    \item Apply the online learning to update the machine learning model during run time.
\end{itemize}

We applied the proposed method to two and three-dimensional multiphase porous media flow problems, as these types of problems represent significant applications in science and engineering, and are very challenging to model due to complex nonlinearities. A great deal of effort has been put in over the years by academia and industry to address these problems~\citep{li:15,salinas:17,jiang:19,freitas:20,silva:21machine,kadeethum2022enhancing}, for which simulation time and computational resources are usually the main constraints. We tested the method in two and three-dimensional cases with very different geometries, such as layers, channel and faults, and also with a considerable range of permeability values. The proposed method can also be applied to other engineering and physical systems as long as they use a numerical relaxation (or other tuning parameter) to control the solver convergence.


The main limitations of the presented technique is the time required to train the machine learning model online and the necessity to calculate the dimensionless numbers in each nonlinear iteration. However, usually the cost of the numerical simulation is by far more expensive than the training operations required, and the mathematical calculations used to compute the dimensionless parameters. Furthermore, we used batch-incremental learning techniques instead of instance-incremental methods in order to minimize the online training cost, and the chosen set of dimensionless numbers reduces the number of calculated features when compared to \citet{silva:21machine}. 

It is worth mentioning that the improvement brought by the online learning (Tables~\ref{tab:rf} and \ref{tab:xgbrf}) is smaller than the offline training (Table \ref{tab:runtime}), although in terms of computational cost it is still relevant. This fact maybe related to the effectiveness of the offline training to learn the optimal relaxation, given the online learning a small room for improvement. In this line, we can also notice that the greatest reductions in the number of nonlinear iterations and walltime (Tables~\ref{tab:rf}, \ref{tab:xgbrf}, and \ref{tab:runtime}) are obtained in test case 4. It is the realistic three-dimensional faulted reservoir model and is the most challenging case from a numerical standpoint, since it has the highest capillary value and comprises several intercalated layers of low and high permeability.

\section{Conclusion}\label{sec:conc}


We propose an online learning acceleration for nonlinear PDE solvers. The proposed approach was applied to complex/realistic two and three-dimentional subsurface reservoirs and was capable of reducing the number of nonlinear iterations without compromising on the accuracy of the results. We also present an analysis of the performance of the presented acceleration technique. This analysis includes the use of different machine learning models, different dimensionless parameters, and different online learning strategies. In this study, we were able to select the most appropriate number of features (dimensionless numbers) used as inputs, and the most suitable machine learning models in terms of accuracy, prediction time and simulation results.
Furthermore, we fully integrated a machine learning model into the nonlinear solver of an open-source, multiphase flow simulator (IC-FERST) and applied it to a set of challenging multiphase porous media flow problems. We were able to reduce simulation time by 37\% on average, and up to 85\% in the best case  (only using the offline training, the online update would further improve the results). 
Compared to other approaches our method is simple to implement and does not require retraining the machine learning model offline when applied to other simulation domains or sets of parameters. Furthermore, not only does it learn offline, but also learns during the simulation to which it is applied. 
We believe that the online/adaptive learning acceleration is not limited to the multiphase porous media flow. It can be applicable to any other system for which a relaxation technique (or other tuning parameter) can be used to stabilise the nonlinear solver.

\section*{Data Availability Statement}

The data, hardware configuration, and source code used in this work are available at \url{https://github.com/viluiz/solver}. 

\section*{Acknowledgements}
We would like to acknowledge financial support from Petrobras for the first author.  This work is also supported by the following EPSRC grants: MUFFINS, MUltiphase Flow-induced Fluid-flexible structure InteractioN in Subsea applications (EP/P033180/1); the PREMIERE programme grant (EP/T000414/1); and SMARTRES, Smart assessment, management and optimisation of urban geothermal resources (NE/X005607/1). 

\section*{Declarations}

\textbf{Competing interests} The authors acknowledge that they have no competing interests associated with this manuscript. 


\bibliography{references}
\bibliographystyle{unsrtnat} 


\clearpage
\appendix

\section{Online training hyperparameters}\label{sec:apphyper}

For the xgboost with the boosting strategy and the random forest with the bagging strategy, we have tested different hyperparameters and the best results were achieved using the values in Tables \ref{tab:hyperxgb} and \ref{tab:hyperrf}, respectively. 

\begin{table}[!htb]
\caption{Hyperparameters used for the online training of the xgboost with the boosting strategy.}
\label{tab:hyperxgb}
\vskip 0.15in
\begin{center}
\begin{small}
\begin{sc}
\begin{tabular}{lc}
\toprule
PARAMETER & VALUE     \\
\midrule
Number of new boosting trees & 1 \\
Learning rate & 0.01 \\
Maximum depth & 3 \\
Subsample of instances & 1.0 \\ 
Subsample of columns & 1.0 \\ 
Buffer size & 50 \\
\bottomrule
\end{tabular}
\end{sc}
\end{small}
\end{center}
\vskip -0.1in
\end{table}

\begin{table}[!htb]
\caption{Hyperparameters used for the online training of the random forest with the bagging strategy.}
\label{tab:hyperrf}
\vskip 0.15in
\begin{center}
\begin{small}
\begin{sc}
\begin{tabular}{lc}
\toprule
PARAMETER & VALUE     \\
\midrule
Number of new trees & 70 \\
Maximum depth & 30 \\
Maximum number of features & 0.2 \\ 
Buffer size & 50 \\
\bottomrule
\end{tabular}
\end{sc}
\end{small}
\end{center}
\vskip -0.1in
\end{table}

\end{document}